\documentclass{article}

\usepackage{arxiv}

\usepackage[utf8]{inputenc} 
\usepackage[T1]{fontenc}    
\usepackage{hyperref}       
\usepackage{url}            
\usepackage{booktabs}       
\usepackage{amsfonts}       
\usepackage{nicefrac}       
\usepackage{microtype}      
\usepackage{lipsum}		
\usepackage{graphicx}
\usepackage{natbib}
\usepackage{doi}
\usepackage{amsmath}
\usepackage{amsthm}
\usepackage{subfigure}

\newtheorem{thm}{Theorem}
\newtheorem{proposition}[thm]{Proposition}

\newtheorem{cor}[thm]{Corollary}
\theoremstyle{definition}

\newtheorem{assumption}{Assumption}
\theoremstyle{remark}

\global\long\def\Expectation{\text{\ensuremath{\mathbb{E}}}}%
\global\long\def\Var{\mathrm{Var}}%
\global\long\def\norm#1{\left\Vert #1\right\Vert }%
\global\long\def\History#1{\mathcal{H}_{#1}}%

\global\long\def\Aspace{\mathcal{A}}%
\global\long\def\Xspace{\mathcal{X}}%
\global\long\def\diag#1{\operatorname{diag}(#1)}%
\global\long\def\partialfrac#1#2{\frac{\partial #1}{\partial #2}}%

\global\long\def\EstName{\texttt{DRUnknown}}

\global\long\def\Qhat{\widehat{Q}}
\global\long\def\Vhat{\widehat{V}}
\global\long\def\phihat{\hat{\phi}}
\global\long\def\betahat{\hat{\beta}}
\global\long\def\muhat{\hat{\mu}}
\global\long\def\rhohat{\hat{\rho}}

\global\long\def\Vtilde{\widetilde{V}}
\global\long\def\Qtilde{\widetilde{Q}}

\global\long\def\etasample{\eta^{(i)}}
\global\long\def\pisample{\pi^{(i)}}
\global\long\def\musample{\mu^{(i)}}
\global\long\def\rhosample{\rho^{(i)}}
\global\long\def\muhatsample{\hat{\mu}^{(i)}}
\global\long\def\rhohatsample{\hat{\rho}^{(i)}}
\global\long\def\Qhatsample{\widehat{Q}^{(i)}}
\global\long\def\Vhatsample{\widehat{V}^{(i)}}

\global\long\def\sumsample{\sum\limits_{i=1}^n}
\global\long\def\sumtime{\sum\limits_{t=0}^{T-1}}
\global\long\def\sumaction{\sum\limits_{a \in \Aspace}}

\title{Doubly-Robust Off-Policy Evaluation with Estimated Logging Policy}

\author{Kyungbok Lee \\
	Graduate School of Data Science\\
	Seoul National University\\
	\texttt{turtle107@snu.ac.kr} \\
	\And
	Myunghee Cho Paik \\\\
	Shepherd23 Inc.\\
	\texttt{myungheechopaik@gmail.com} \\
}

\renewcommand{\shorttitle}{Doubly-Robust Off-Policy Evaluation with Estimated Logging Policy}

\hypersetup{
pdftitle={Doubly-Robust Off-Policy Evaluation with Estimated Logging Policy},
pdfsubject={statistics, machine learning},
pdfauthor={Kyungbok Lee, Myunghee Cho Paik},
pdfkeywords={Off-policy evaluation, Doubly robust},
}

\begin{document}
\maketitle

\begin{abstract}
We introduce a novel doubly-robust (DR) off-policy evaluation (OPE) estimator for Markov decision processes, \EstName{}, designed for situations where both the logging policy and the value function are unknown.
The proposed estimator initially estimates the logging policy and then estimates the value function model by minimizing the asymptotic variance of the estimator while considering the estimating effect of the logging policy. When the logging policy model is correctly specified, \EstName{} achieves the smallest asymptotic variance within the class containing existing OPE estimators. 
When the value function model is also correctly specified, \EstName{} is optimal as its asymptotic variance reaches the semiparametric lower bound.
We present experimental results conducted in contextual bandits and reinforcement learning to compare the performance of \EstName{} with that of existing methods.
\end{abstract}


\section{Introduction}

In various decision-making problems, estimating the value, the expected reward of a policy is a crucial question that needs to be addressed. Online evaluation requiring a comprehensive evaluation of policy value can be expensive and may not be applicable to multiple target policies.

Alternatively, \textit{off-policy} evaluation (OPE) refers to a technique that estimates the value of a \textit{target} policy by utilizing log data generated from a different \textit{logging} policy. This approach has attracted considerable interest in the domains of  contextual bandits (CB) \citep{dudik2011doubly, swaminathan2017off} and reinforcement learning (RL) \citep{precup2000eligibility, mahmood2014weighted, jiang2016doubly}.

Several off-policy evaluation algorithms \citep{dudik2011doubly, thomas2016data, wang2017optimal, farajtabar2018more, su2020doubly} currently in use rely on having complete knowledge of the logging policy in order to utilize inverse probability weighting (IPW). However, in situations where information about the data logger is not available, such as when using human decision data, it becomes necessary to estimate the logging policy. Applying existing estimators directly to cases where the logging policy is unknown can compromise the desired properties of these methods, as they fail to take into account the impact of logging policy estimation.

In this paper, we present a novel efficient estimator called \EstName{}, which simultaneously estimate both the logging policy model and the value function model. Our proposed approach, without the need for external data, effectively captures the interdependency between the two models.

In the field of conventional statistics, \citet{cao2009improving} employs the influence function to estimate parameters in the doubly-robust (DR) estimator for population means when there are missing observations.
The study focused on situations where data is incomplete and the missing mechanisms are unknown and estimated.
To advance this concept, we utilized the influence function approach to address the OPE problem. Here, we treat the rewards of unselected actions as missing values and estimate the unknown logging policy. This enables us to derive the asymptotic variance of the OPE estimator and estimate the parameters that minimize it.

The proposed OPE estimator estimates both the logging policy model and the value function model, and is referred to as \textit{doubly-robust} due to its consistency if either the model for the logging policy or the value function is correctly specified.
When the logging policy is correctly specified, the proposed estimator is the most efficient among the class of DR OPE estimators with an estimated logging policy, and is at least as efficient as existing methods.
Moreover, if the value function model is correctly specified, the estimator  locally efficient, achieving the semiparametric lower bound for asymptotic variance and is asymptotically optimal.
In order to demonstrate the effectiveness of the proposed estimator, we conducted simulation experiments to compare its performance with previous methods in contextual bandits and reinforcement learning problems.

The main contributions of this paper are as follows:
\begin{itemize}
    \item We present a new DR OPE estimator for Markov decision process, for the case where both the logging policy and the value function are unknown. The proposed estimator is consistent when either the logging policy model or the value function model is correctly specified.
    
    \item We propose a method to estimate both the logging policy model parameter $\phihat$ and the value function model parameter $\betahat$ simultaneously. We use the maximum likelihood estimator (MLE) for $\phihat$ and minimize the asymptotic variance to estimate $\betahat$, accounting for the impact of estimating $\phihat$.

    \item The proposed estimator has the smallest asymptotic variance among estimators using MLE for $\phihat$ when the logging policy model is correctly specified. When the value function model is also correctly specified the proposed estimator is optimal, as its asymptotic variance reaches the semiparametric lower bound.
    
    \item Simulations on contextual bandits and reinforcement learning problems show that the proposed estimator consistently shows smaller mean-squared errors compared to the benchmarks methods.
\end{itemize}

\section{Problem Setup}
In this paper, we model the decision-making problem as a reinforcement learning framework in which the learner's interaction with the system is represented as a \textit{Markov Decision Process} (MDP). In this section, we provide definitions of MDP and the corresponding off-policy evaluation problem for our study.

\subsection{Markov Decision Processes}\label{subsec:MDP}
An MDP is defined as a tuple ($\Xspace$, $\Aspace$, $R, P, P_0, \gamma$), where $\Xspace$ and $\Aspace$ represent the state and the finite action spaces with $|\Aspace|=K$, $R(x, a)$ denotes the distribution of the random variable $r(x, a)$ of the bounded immediate reward of taking action $a$ in state $x$, $P(\cdot|x, a)$ is the transition probability distribution. $P_0 : X \rightarrow [0, 1]$ is the initial state distribution, and $\gamma \in [0, 1]$ is the discounting factor.
The learner utilizes a policy $\pi : \Xspace \times \Aspace \rightarrow [0, 1]$, a stochastic mapping from states to actions. Here, $\pi(a|x)$ represents the probability of taking action $a$ in state $x$.

Let $\History{T}$ = $\{(x_t, a_t, r_t)\} _{t=0}^{T-1}$ represent a $T$-step trajectory containing state-action-rewards generated by policy $\pi$.
Denote $R_{t_1:t_2} = \sum\limits_{t=t_1}^{t_2} \gamma^{t-t_1} r_t$ as the sum of discounted return of from time step $t_1$ to $t_2$. 
Our goal is to estimate the \textit{policy value}, the expected return of trajectories generated by policy $\pi$, i.e., $V^{\pi,T}= \Expectation_{\pi}  [R_{0:T-1}]$. Here $\Expectation _{\pi}$ denotes the expectation over the trajectory sampled from $\pi$. 
We also define the value function of a policy $\pi$ at state $x$ and at state-action pair $(x,a)$ from time step $t$ by $V^{\pi,t}(x)=\Expectation_{\pi}  [R_{t:T-1}|x_t = x]$ and $Q^{\pi,t}(x,a)=\Expectation_{\pi}  [R_{t:T-1}|x_t = x, a_t=a]$, respectively.

Throughout this paper, we fix the trajectory length as $T$. We also omit the $\pi$ from $V^{\pi,t}(s)$ and $Q^{\pi,t}(s,a)$ to simplify them as $V^t(s)$ and $Q^t(s,a)$, respectively, since we are only interested in the value of $\pi$. Note that, unlike the case with $T=\infty$, where the $Q$ and $V$ value functions remain unchanged, the value functions $Q^{t}$ and $V^{t}$ can vary for each time step $t$ in general.

The contextual bandits (CB) is a special case of the MDP, in which the trajectory length $T$ equals to $1$, and the transition dynamics $P$ do not exist. In CB, the state is represented as the context, and the action is called as the arm.

\subsection{Off-Policy Evaluation Problem}

In an OPE problem, we are given a dataset $\mathcal{D}_n = \{\History{T}^{(i)}\}_{i=1}^n$ comprising $n$ independent $T$-step trajectories drawn from a logging policy $\mu$. We use the superscript $(i)$ to refer to the data from the $i^{\text{th}}$ trajectory, denoted as $\History{T}^{(i)}$. The objective is to estimate the policy value $V^{\pi}$ of a separate target policy $\pi$.
We define the \textit{importance ratio} $\rho_{t}=\pi(a_t|x_t) / \mu(a_t|x_t)$, and the \textit{cumulative importance ratio} from time step $t_1$ to $t_2$ between $\mu$ and $\pi$ as
$\rho_{t_1: t_2}= \Pi_{t=t_1}^ {t_2} \rho_t$. In cases where $t_1 > t_2$, we define $\rho_{t_1:t_2}$ to be 1.
Similarly, we denote the estimated importance ratio as $\rhohat$, where the probability of the logging policy $\mu(a|x)$ is replaced with its estimated version $\muhat(a|x;\phihat)$ for the logging policy model $\muhat(\cdot;\phi)$.

The policy value $\Vhat$ is a point estimator, and there are various metrics to assess the quality of the provided estimator. One commonly used is the \textit{mean squared error} (MSE), defined as
$\text{MSE}(\Vhat) = \Expectation_{\mu}[{(\Vhat - V^{\pi})^2}]$
with the expectation taken with respect to the data sampled from $\mu$.
MSE quantifies the performance of a given estimator in finite samples. Another measure is the \textit{asymptotic variance}, which assesses the efficiency of an estimator with a large number of samples. The estimator $\Vhat$ is called \textit{asymptotically linear} if there exists a function $\psi$ of trajectory $\History{T}$ such that
$$
\Vhat - V^{\pi} = \frac{1}{n}\sum_{i=1}^n \psi(\History{T}^{(i)}) + o_p(\frac{1}{\sqrt{n}}).
$$
with $\Expectation_{\mu}{[\psi(\History{T})]}=0$ and $\Var_{\mu}{[\psi(\History{T})^2]} < \infty$. The quantity $\Var_{\mu}{[\psi(\History{T})^2]}$ is denoted as the asymptotic variance of $\Vhat$, and the function $\psi(\History{T})$ is referred to as the \textit{influence function}. In our study, we derive the specific form of the influence function for OPE estimators and conduct an analysis to minimize the asymptotic variance.

For our analysis, we use the following standard regularity assumptions for OPE problem.
\begin{assumption}[Absolute Continuity]\label{assumption:AC}\ \\
For all state-action pair $(x,a) \in \Xspace \times \Aspace$, if $\pi(a|x)>0$ then $\mu(a|x)>0$.
\end{assumption}
\begin{assumption}[Square Integrable]\label{assumption:integrable}\ \\
$\Expectation_{\mu}{[\rho_{0:t}^2]}<\infty$ for all $t<T$. 
\end{assumption}
Assumption \ref{assumption:AC} is crucial in order to avoid an infinite IPW and the OPE estimators which employ IPW are well-defined. Assumption \ref{assumption:integrable} is necessary to guarantee that estimators utilizing the IPW method have finite variance.

\section{Previous Methods for Off-Policy Evaluation}
The doubly-robust type estimator for the OPE problem  was first proposed by \citet{dudik2011doubly}. The estimator depends on a regression model for value function that is estimated using a separate independent dataset, which may not always be available. 

The works of \citet{thomas2016data} and \citet{wang2017optimal} introduce estimators that improve efficiency by modifying the IPW component of the DR OPE estimator. These methods still require the independently estimated value function using external data. 
\citet{farajtabar2018more} employed the idea of \citet{rubin2008empirical}, which estimates the value function model by minimizing the estimation variance without external dataset. All these methods assume that the true logging policy is known and do not require estimation.

Several studies \citep{li2015toward, raghu2018behaviour, DBLP:conf/iclr/XieLLWZP19, hanna2021importance} have proposed the IPW estimators for OPE utilizing the estimated logging policy and examined their theoretical and numerical properties of these estimators. However, these methods do not consider the value function, making them not doubly-robust and suboptimal in terms of asymptotic variance.

\subsection{Standard Methods for Off-Policy Evaluation}
There are three standard approaches to estimate $V^{\pi}$ in OPE problem, and we provide a brief overview of these methods.

The first method is \textit{direct method} (DM) \citep{dudik2011doubly, rothe2016value}, which directly approximates the state-action value function $Q(x,a)$ of the target policy. The DM does not use the information from the logging policy $\mu$ and relies heavily on the accuracy of the prediction of the value function. While the DM estimators often exhibits low variance, it is not guaranteed to be unbiased.

The second method is called \textit{inverse probability weighting} (IPW) estimator \citep{horvitz1952generalization}, which utilize the importance ratio term to correct the discrepancy between the target and logging policies. The IPW estimator is unbiased in case the logging policy is known, but it can exhibit a large variance when there is a significant difference between the logging and target policies and the importance ratio have large value. The Assumption 1 is required in order for the IPW estimator to be well-defined: one cannot know about the pair $(x,a)$ never explored by the logging policy.
 
The third approach, named \textit{doubly-robust} (DR) estimator \citep{cassel1976some, robins1995semiparametric, dudik2011doubly, jiang2016doubly}, combines the DM and IPW, leveraging their respective strengths to obtain favorable characteristics. The IPW estimator is a special case of DR with the value function fixed to zero. Our work focuses on the DR type estimators, as our main objective is to introduce an efficient method for simultaneously estimating the logging policy and the value function model.

\subsection{Doubly-Robust Off-Policy Evaluation with Unknown Logging Policy}

The standard DR OPE estimator with known logging policy is given by
\begin{equation}
\Vhat^{\text{DR}} = \frac{1}{n} \sumsample \sumtime \gamma^{t} \rho_{0:{t-1}}^{(i)}
\bigl[\rho_t^{(i)}[r_t^{(i)} - \Qhat(x_t^{(i)}, a_t^{(i)};\beta)] + \Vhat(x_t^{(i)};\beta)\bigr],       
\end{equation}\label{eq:DR}
 
where $\Qhat(\cdot ;\beta)$ is a \textit{value function model} of the state-action value function parameterized by $\beta$,
and $\Vhat(x ;\beta) = \sum_{a \in \Aspace} \pi(a|x) \Qhat(x,a ;\beta).$ 

Here, $\Qhat$ depends solely on the state-action pair. However, as mentioned in Section \ref{subsec:MDP}, the $Q^t(x,a)$ may vary for different time steps, even if the state-action pair remains the same when the horizon length $T$ is finite. Hence, we cannot ensure that the model $\Qhat$ can express the $Q^t$ for all $t$ in the OPE problem with a finite $T$.
To address this issue, we utilize the function class $\Qhat(t,x,a)$, which incorporates $t$ as an input parameter in practice.
Also, to ensure that the class of DR OPE estimator contains the IPW estimator, we assume that the class of $\Qhat$ contains the constant functions. This can be easily satisfied by bringing the parameters $\nu_0, \nu_1 \in \mathbb{R}$ to extend the function class by $\nu_0 + \nu_1 \Qhat$.

The DR OPE estimator \eqref{eq:DR} utilizes the importance ratio using the true probability of the logging policy.
When the logging policy $\mu$ is unknown we use the logging policy model $\muhat(\cdot ;\phihat)$ to approximate the unknown $\mu$.
We say that the model $\muhat$ is \textit{correctly specified} if there exists a value of $\phi$ such that $\muhat(x,a;\phi)=\mu(x,a)$ for all $x$ and $a$. 
The correct specification of $\muhat$ is necessary for the consistent estimation of the policy value $V^{\pi}$, as correctly specifying the value function model $\Qhat$ is difficult in many OPE problems.

In the following sections, our focus will be on the DR OPE estimator, which estimates both the logging policy parameter $\phi$ and the value function parameter $\beta$. Our goal is to build an efficient OPE estimator that exhibits lower asymptotic variance compared to existing methods.

\section{Proposed Method: \EstName{}}
In this section, we present our class of \EstName{} estimators. The central concept of \EstName{} is to first estimate the unknown logging policy parameter $\phihat$ for the IPW component of the DR estimator, and then learn the value function parameter $\betahat$ to minimize the asymptotic variance of the estimator.
We utilize the influence function of the estimator to derive a feasible objective function to minimize the asymptotic variance of \EstName{}.

\subsection{\EstName{} for Contextual Bandits}\label{subsec:DRUnknown_CB}
We first present the \EstName{} for contextual bandits problem with $T=1$. Consider a logging policy model $\muhat(x,a ; \phi)$ parameterized by $\phi$.
Let $\Delta_{a}^{(i)}$ represent the indicator for an action $a$ chosen in trajectory $i$, and denote the partial derivative of $\muhat$ with respect to $\phi$ as $\dot{\muhat} (a|x; \phi)$. The \textit{maximum-likelihood estimator} (MLE) $\phihat = \phihat_n$ is given by the solution of the estimating equation $U_n(\phi)$ given by
\begin{equation*}
U_n(\phi)=\sumsample \sumaction \Delta_{a}^{(i)} \frac{\dot{\muhat} (a|x^{(i)}; \phi)}{\muhat (a|x^{(i)}; \phi)}=0. 
\end{equation*}
If the logging policy model is correctly specified, the estimating equation is unbiased, and $\phihat$ is consistent for $\phi$.

Plugging in the estimated value of $\phihat$, we obtain the following class of DR OPE estimator
$$
\Vhat^{\text{DR}}(\beta, \phihat)
= \frac{1}{n} \sumsample \Bigl[ \frac{\pisample}{\muhatsample}
(r^{(i)} - \Qhatsample(\beta)) + \Vhatsample(\beta) \Bigr]   
$$
where the notations $\pisample, \muhatsample, \Qhatsample(\beta)$ and $\Vhatsample(\beta)$ denote the $\pi(a^{(i)}|x^{(i)})$, $\muhat(a^{(i)}|x^{(i)};\phihat)$, $\Qhat(x^{(i)}, a^{(i)}; \beta)$ and $\Vhat(x^{(i)}; \beta)$ respectively. Henceforth in the paper, if there is no ambiguity, we simplify terms containing $x^{(i)}$ and $a^{(i)}$ by using only the superscript $(i)$.

The expression of the bias and variance of the estimator $\Vhat^{\text{DR}}(\beta, \phihat)$ is not simple in general, even for a fixed $\beta$. The estimator is not a mean of the independent and identically distributed (i.i.d) variables, due to the estimated $\phihat$ in the denominator. Consequently, estimating the regression parameter $\beta$ to minimize the MSE of the proposed estimator poses challenges. Hence, we instead aim to find $\betahat$ minimizing the asymptotic variance of the estimator.

To attain this, we derive the influence function asymptotically equivalent to $\Vhat$, through a Taylor expansion with respect to $\phi$ and $\beta$.
Denote the new regression function $F$, the product of the target policy $\pi$ and the $\Qhat$ with an additional estimating effect term of $\phi$, given by
$$
F(x,a;\beta, c, \phi)=\pi(a|x)\Qhat(x,a;\beta) + c^{\top} \dot{\muhat}(a|x;\phi), 
$$
for $\beta$ and $c \in \mathbb{R}^{\text{dim}(\phi)}.$
The influence function of $\Vhat^{\text{DR}}(\betahat,\phihat)$ with arbitrary estimator $\betahat$ can be formulated employing $F$, as described in the following proposition.
 
\begin{proposition}[Asymptotic Equivalence of $\Vhat$ for CB]\label{prop:influence}
 Consider an estimator $\betahat$ converging to some $\beta^*$ in probability. If the model $\muhat(\cdot;\phi)$ is correctly specified, the DR OPE estimator is asymptotically linear with influence function $\eta$:
\begin{equation*}
\Vhat^{\text{DR}}(\betahat, \phihat) = \Vtilde(\beta^*, c(\beta^*)) + o_p(n^{-1/2})
= \frac{1}{n} \sumsample \etasample(\beta,c(\beta^*)) + o_p(n^{-1/2}),
\end{equation*}
where $c(\beta)$ is a vector that solely depends on $\beta$, and
\begin{equation*}
\etasample(\beta,c)=\frac{1}{\musample}\bigl[\pisample r^{(i)} -F^{(i)}(\beta,c,\phi) \bigr] \\
+\sumaction F(x^{(i)},a; \beta, c,\phi).    
\end{equation*}\label{eq:true_IF}
\end{proposition}
As the value of the vector $c(\beta^*)$ is unknown, we estimate both $\beta$ and $c$ that minimize the variance of $\Vtilde(\beta,c)$.
Denote the vector $\vec{F}(x; \beta,c,\phi)=(F(x, a; \beta,c,\phi))_{a \in \Aspace}$ and $\pi\vec{Q}(x) = (\pi(a|x) Q(x,a))_{a \in \Aspace}$, and the gradient matrix
$f(x;\beta,c,\phi) =\displaystyle\frac{\partial \vec{F}}{\partial(\beta,c)}.$
By the law of total variance, the variance of $\Vtilde^{\text{DR}}(\beta,c)$ can be expressed as the square of stochastic seminorm added by a constant independent of $\beta$ and $c$.
\begin{proposition}[Variance of $\Vtilde^{\text{DR}}$]\label{prop:bandit_variance}
The variance of  $\Vtilde(\beta,c)$ is given by
$$
n\Var{(\Vtilde(\beta,c))} = \Var (V^{\pi}(x)) + \Expectation_{\mu}\displaystyle\norm{\pi\vec{r} - \pi\vec{Q}(x) }_{M_{\mu}}^2 
+ \Expectation_{\mu}\displaystyle\norm{\vec{F}(x;\beta,c,\phi)-\pi\vec{Q}(x) }_{M_{\mu}}^2   
$$
where $M_{\mu}=\diag{\mu(a|x)^{-1}}_{a \in \Aspace} - J_K$ and $J_K \in \mathbb{R}^{K \times K}$ the matrix of ones.
\end{proposition}

The Proposition \ref{prop:bandit_variance} tells minimizing the square of seminorm
\begin{equation}\label{eq:bandit_seminorm}
\Expectation_{\mu}{\norm{\vec{F}(x;\beta,c,\phi)- \pi \vec{Q}(x)}_{M_{\mu}}^2}
\end{equation}
with respect to $\beta$ and $c$ is equivalent to minimizing the variance of the $\Vtilde(\beta,c)$.
Denote the minimizer of \eqref{eq:bandit_seminorm} by $(\beta_{\text{opt}}, c_{\text{opt}})$, the zero of its gradient
\begin{equation*}\label{eq:bandit_gradient}
\begin{split}
\Expectation_{\mu} \bigl[ f^{\top}(x;\beta,c,\phi) M_{\mu} (\vec{F}(x;\beta,c,\phi)- \pi \vec{Q}(x)) \bigr]= 0.
\end{split}
\end{equation*}
We can observe that the solution satisfies $c_{\text{opt}}=c(\beta_{\text{opt}})$, so that the variance of $\Vtilde(\beta_{\text{opt}},c(\beta_{\text{opt}}))=\Vtilde(\beta_{\text{opt}},c_{\text{opt}})$ is minimized at $(\beta_{\text{opt}}, c_{\text{opt}})$. Therefore by Proposition \ref{prop:influence}, the smallest asymptotic variance of DR OPE estimator $\Vhat^{\text{DR}}(\betahat,\phihat)$ under the correctly specified $\muhat$, is achieved by $\betahat$ converging in probability to this $\beta_{\text{opt}}$.The following estimating equation $S_n(\beta,c)$ is a tractable equation that jointly solves for $(\beta,c)$, with its solution $\betahat$ consistent for $\beta_{\text{opt}}$:
\begin{equation}\label{eq:bandit_gradient_data}
S_n(\beta,c)= \sumsample f^{(i) \top}(\beta,c)
M^{(i)} \bigl( F^{(i)}(\beta,c) - \diag{\pi(a|x^{(i)})}_{a \in \Aspace} ~ \vec{r}^{(i)}\bigr)= 0
\end{equation}
for $M^{(i)}=\displaystyle\diag{\muhat(a|x^{(i)})^{-1}}_{a \in \Aspace} - J_K$ and the pseudo reward
$\vec{r}^{(i)} = \displaystyle\bigl(\frac{\Delta_a^{(i)}}{\muhat(a|x^{(i)})} r^{(i)}\bigr)_{a \in \Aspace}$.
Plugging in the solution $\betahat$ of $S_n(\beta,c)=0$ as $\Vhat^{\text{DR}}(\betahat, \phihat)$, we obtain our proposed estimator, \EstName{}. The $c$ serves as an auxiliary parameter that adjusts for the effect of estimating $\phihat$, and does not have a direct role in the final estimator $\Vhat^{\text{DR}}(\betahat, \phihat)$.

The proposed \EstName{} is doubly-robust within the statistical context, as it remains consistent if either the logging policy model $\muhat$ or the value function model $\Qhat$ is correctly specified. 
We have addressed the scenario with the correct logging policy model above. Regarding the second scenario, the minimizer of \eqref{eq:bandit_seminorm} is $(\beta_{\text{opt}},c_{\text{opt}})=(\beta_0, 0)$ since $\Gamma(\beta_0)=0$, where $\beta_0$ is the true value of the parameter satisfying $Q(\cdot)=\Qhat(\cdot;\beta_0).$ Consequently, the solution of \eqref{eq:bandit_gradient_data} converges to $(\beta_0, 0)$, establishing the consistency of \EstName{}. This observation can be summarized as the following proposition.
\begin{proposition}[Doubly-Robustness]\label{prop:DR_property_bandit}
The \EstName{} $\Vhat^{\text{DR}}(\betahat, \phihat)$ for contextual bandits is a doubly robust: it converges to $V^{\pi}$ in probability if either the logging policy model $\muhat$ or the value function model $\Qhat$ is correctly specified.
\end{proposition}

\subsubsection{Intuitive Understanding of \EstName{} for CB: extended function class}
For an intuitive understanding of \EstName{} for contextual bandits, suppose that the target policy $\pi$ always have a nonzero value. Then, $F$ can be rewritten as
$
F(x,a;\beta,\phi) = \pi(a|x)\Qtilde(x,a;\beta,c)
$
for the extended function class 
$$\Qtilde(x,a;\beta,c)=\Qhat(x,a;\beta,c) + c^{\top} \displaystyle\frac{\dot{\muhat}(a|x;\phi)}{\pi(a|x)}.$$
The objective function \eqref{eq:bandit_seminorm} we aim to minimize in \EstName{} can then be expressed as
$$
\Expectation_{\mu}{\norm{\bigl( \pi(a|x)[\Qtilde(x,a;\beta,c) - Q(x,a)] \bigr)_{a \in \Aspace} }_{M_{\mu}}^2}.
$$
The additional linear term  $\displaystyle c^{\top}\frac{\dot{\muhat}(a|x;\phi)}{\pi(a|x)}$ from $\Qtilde$ can be seen as the projection of $Q^{\pi}$ onto the inverse probability tangent space, the linear space spanned by the score function of $\phihat$.

Hence, the proposed estimator can be seen as employing a linear parameter $c$ within the expanded function class $\Qtilde$ to remove the effect of estimating $\phihat$, while seeking the optimal $\betahat$ for $\Qhat$ that minimizes the asymptotic variance.

\subsection{\EstName{} for Reinforcement Learning}
The proposed \EstName{} for RL is constructed similarly to the case of CB. However, its analysis is more complex, as the estimator is expressed as a weighted sum of terms observed from time step $0$ to $T-1$. 
Below, we introduce the construction of \EstName{} for RL, commencing with the estimation of $\hat{\phi}$.

The MLE $\phihat_n = \phihat$ of $\phi$ for RL is given by the solution of 
\begin{equation*}
 U_n(\phi)=\sumsample \sumtime \sumaction \Delta_{a,t}^{(i)} \frac{\dot{\muhat}(a|x_t^{(i)}; \phi)}{\muhat (a|x_t^{(i)}; \phi)}=0,   
\end{equation*}
where $\Delta_{a,t}^{(i)}$ is the indicator variable for selected action at time step $t$ from trajectory $i$.  
Substituting the MLE $\phihat$ into the DR estimator \eqref{eq:DR}, the estimator is given by
$\Vhat^{\text{DR}}(\beta,\phihat) = \displaystyle\sumtime \gamma^{t} \Vhat_{t}(\beta,\phihat),$
with
$$
\Vhat_{t}(\beta,\phihat) = \frac{1}{n} \sumsample  \rhohat_{0:{t-1}}^{(i)}
\bigl[\rhohat_t^{(i)}[r_t^{(i)} - \Qhatsample(\beta)] + \Vhatsample(\beta)\bigr].
$$
The estimator of the value at each time step $t$, $\Vhat_t(\beta,\hat{\phi})$, takes the same form of \EstName{} for CB, weighted by $\rhohatsample_{0:{t-1}}$. However, additional analysis is needed in the case of RL, as the $\Vhat_t$ for each $t$ are stochastically correlated. Now, as done in CB, we define the value regression function $F_t(x,a;\beta,c,\phi)$ for each $t$, given by
$$
F_{t}(x,a;\beta, c,\phi)
=\rho_{0:t-1}\pi(a|x)\Qhat(x,a;\beta) + \gamma^{-t} c^{\top} \dot{\muhat}(a|x;\phi).    
$$
and we derive the influence function of the proposed estimator as stated in the following proposition. 
\begin{proposition}[Asymptotic Equivalence of $\Vhat$ for RL]\label{prop:influence_RL}
 Consider an estimator $\betahat$ converging to some $\beta^*$ in probability. If the logging policy model $\muhat(\cdot;\phi)$ is correctly specified, then the proposed DR OPE estimator is asymptotically linear, with influence function $\eta=\displaystyle\sumtime\gamma^{t}\eta_t$:
\begin{equation}
\Vhat^{\text{DR}}(\betahat, \phihat) = \Vtilde(\beta^*, c(\beta^*)) + o_p(n^{-1/2}) = \frac{1}{n} \sumsample \etasample(\beta,c(\beta^*)) + o_p(n^{-1/2}),
\end{equation}
where $c(\beta)$ is a vector that solely depends on $\beta$, and
$$
\eta_t^{(i)}(\beta,c) = \frac{1}{\musample_t}\bigl[ \rho_{0:t-1}^{(i)} \pisample_t r_t^{(i)} -F_{t}^{(i)}(\beta,c,\phi) \bigr]
+\sumaction F_{t}(x_t^{(i)},a; \beta, c,\phi).
$$
\end{proposition}

Now, to derive the asymptotic variance of the proposed estimator for RL, we denote the vector
$\vec{F}_{t}(x; \beta,c, \phi)=(F_t(x, a; \beta,c, \phi))_{a \in \Aspace},$
$\pi \vec{Q}^t(x) = (\pi(a|x) Q^t(x,a))_{a \in \Aspace}$ and the gradient matrix
$ f_t(x;\beta,c, \phi) =\displaystyle\frac{\partial \vec{F}_t}{\partial(\beta,c)}.$

The asymptotic variance of the DR OPE estimator for RL can be expressed with these notations as in the following proposition.
\begin{proposition}[Variance of $\Vtilde$ for RL]\label{prop:RL_variance}
The variance of  $\Vtilde(\beta,c)$ is given by
$$
n\Var{(\Vtilde(\beta,c))} = C_T
 +\sumtime \gamma^{2t} \Expectation_{\mu}  \norm{\vec{F}_{t}(x_t;\beta,c,\phi)-  \rho_{0:t-1} \pi \vec{Q}^t(x_t)}^2_{M_{t,\mu}}
$$
where $M_{t,\mu}=\diag{\mu(a|x_t)^{-1}}_{a \in \Aspace} - J_K$ and
$$C_T = \sumtime \gamma^{2t} \Expectation_{\mu} \bigl[ \Var_t \left[ \rho_{0:t-1}V^t(x_t) \right] + \Var_{t+1}[\rho_{0:t}r_t] \bigr],$$
with $\Var_{t}$ the variance conditioned on the history up to time step $t-1$, $\{x_0,a_0,\dots, x_{t-1},a_{t-1}\}$.
\end{proposition}

Similar to the case of CB, the proposition implies that minimizing the weighted sum of seminorms with respect to $\beta$ and $c$ is equivalent to minimizing the asymptotic variance of $\Vhat^{\text{DR}}$. The solution $\hat{\beta}$ of the following estimating equation is consistent for $\beta_{\text{opt}}$, minimizing the asymptotic variance of $\Vhat^{\text{DR}}$ by $\sigma^2(\beta^*)=\Var_{\mu}{[\eta(\beta^*, c(\beta^*))]}$.
\begin{equation}\label{eq:RL_gradient_data}
S_n(\beta,c)=\sumsample \sumtime \gamma^{2t} f_{t}^{(i) \top}(\beta,c,\phihat)
M_t^{(i)}\bigl(\vec{F}_{t}^{(i)}(\beta,c) - \rhohat_{0:t-1} \diag{\pi_t^{(i)}}_{a \in \Aspace} ~ \vec{r}_{t:T-1}^{(i)}\bigr)= 0,
\end{equation}
with the pseudo-reward
$\displaystyle{\vec{r}^{(i)}_{t:T-1}} = \bigl( \frac{\Delta_{a,t}^{(i)}}{\muhat(a|x_t^{(i)})}  \bar{R}_{t:T-1}^{(i)} \bigr)_{a \in \Aspace}$
for
$\bar{R}_{t:T-1}^{(i)} = r_t^{(i)} + \displaystyle\sum\limits_{\tau=t+1}^{T-1} \gamma^{\tau-t} \rhohat_{t+1:\tau}^{(i)} r_{\tau}^{(i)},$
and $M_t^{(i)}=\displaystyle\diag{\muhat(a|x_t^{(i)})^{-1}}_{a \in \Aspace} - J_K$.

\section{Theoretical Properties}
We now present the theoretical properties of the proposed estimator, with a focus on its asymptotic distribution. Combining Proposition \ref{prop:influence}, \ref{prop:bandit_variance}, \ref{prop:influence_RL} and \ref{prop:RL_variance}, the \EstName{} has the following asymptotic normality for both contextual bandits and reinforcement learning problems.

\begin{thm}[Asymptotic distribution]
    When the logging policy model $\muhat$ is correctly specified, the proposed \EstName{} estimator is asymptotically normal, given by 
    $$\sqrt{n}(\Vhat^{\text{DR}}(\betahat,\phihat) - V^{\pi}) \xrightarrow{d} \mathcal{N} (0, \sigma^2)$$
    for $\sigma^2=\sigma^2(\beta^*).$
\end{thm}
To calculate the confidence interval of $V^{\pi}$, one needs to estimate the unknown variance $\sigma^2$. As we know that
\begin{equation*}
\begin{split}
\sigma^2 &= \sigma^2(\beta^*, c(\beta^*), \phi) \\
&= \Var_{\mu}{[\Vtilde(\beta^*,c(\beta^*))]} = n \Var_{\mu}[\eta(\beta^*,c(\beta^*))],        
\end{split}
\end{equation*}
we have
$\hat{\sigma}^2 (\beta, c(\beta^*), \phi) = \displaystyle\frac{1}{n} \sumsample (\etasample(\betahat,\hat{c},\phihat) - \bar{\eta})^2$
is a consistent estimator for $\sigma^2$, with
$\bar{\eta}=\displaystyle\frac{1}{n}\sumsample \etasample(\betahat,\hat{c},\phihat)$
the average of $\etasample$ and $\betahat, \hat{c}$ are the solutions of \eqref{eq:RL_gradient_data} and the MLE $\phihat$. By the Slutzky's theorem, we have
$$
\hat{\sigma}^2 = \hat{\sigma}^2(\betahat, \hat{c}, \phihat) \xrightarrow{p} \sigma^2,
$$
and combining the result with the theorem we have the $(1-\alpha)$ confidence interval as
$$
\bigl[ \Vhat^{\text{DR}}(\betahat,\phihat) - z_{\alpha/2} \frac{\hat{\sigma}^2}{\sqrt{n}}, \Vhat^{\text{DR}}(\betahat,\phihat) + z_{\alpha/2} \frac{\hat{\sigma}^2}{\sqrt{n}}  \bigr],
$$
with $z_{\alpha/2}$ the $\alpha/2$ standard Gaussian quantile.

When the value function model $\Qhat$ is also correctly specified, the proposed \EstName{} is asymptotically equivalent to the DR OPE estimator using the true logging policy and the value function, and is locally efficient.
\begin{proposition}[Local Efficiency]\label{prop:local_efficiency}
When both the logging policy model $\muhat$ and the value function model $\Qhat$ are correctly specified, 
The asymptotic variance of the proposed estimator achieves the semiparametric lower bound and is asymptotically optimal.
\end{proposition}
Proposition \ref{prop:local_efficiency} states that \EstName{} can be asymptotically optimal when the model $\widehat{Q}$ is properly chosen. Also, for any choice of value function model, whether it is correct or not, the proposed estimator is asymptotically more efficient compared to existing OPE algorithms, including IPW, DR, and MRDR, in the OPE problem with an unknown logging policy.

\begin{proposition}[Intrinsic Efficiency]\label{prop:intrinsic_efficiency}
    The proposed DR OPE estimator has the smallest asymptotic variance among the class of DR OPE using MLE for $\phi$, when the logging policy model $\muhat$ is correctly specified. The same holds for arbitrary consistent estimator of $\phi$.
\end{proposition}

\begin{cor}[Comparison to Existing Algorithms]
The proposed estimator has at least smaller asymptotic variance than IPW, DR and MRDR utilizing the same estimator for $\phi$, when the logging policy model $\muhat$ is correctly specified.
\end{cor}

\section{Experimental Results}
In this section, we compare the performance of the four estimators: (i) IPW \citep{horvitz1952generalization} (ii) MLIPW \citep{DBLP:conf/iclr/XieLLWZP19} (iii) MRDR \citep{farajtabar2018more} and (iv) the proposed \EstName{} on CB and RL problems.
IPW requires knowledge of the true logging policy $\mu$ and cannot be applied in our experimental scenario. Therefore it serves as a baseline for the other three estimators, and we compare the relative MSE of each estimator.

\subsection{Contextual Bandits}
\subsubsection{Simulation Data}

We use the simulation environments described as follows. We generate $N=100$ test datasets with size $n$, given by $D_{j} = \{(x_{i}^{j}, a_{i}^{j}, r_{i}^{j})\}_{i=1}^n,~j \in [N]$. The context $x_{i}^{j}$ for $i^{\text{th}}$ sample from $j^{\text{th}}$ dataset contains $d=5$ dimensional $K=10$ vectors randomly sampled from uniform distribution $U(-1/\sqrt{d},1/\sqrt{d})$. The rewards are generated from the Gaussian distribution with mean $\exp (x_{i}^{j \top} \beta)$ and variance 1, where $\beta$ is a fixed coefficient also sampled from the uniform distribution $U(-1/\sqrt{d},1/\sqrt{d})$.

The logging policy $\mu$ and target policy $\pi$  follows the linear logistic model with random coefficients $\phi_{\mu}$ and $\phi_{\pi}$ as
$\mu(a|x) = \displaystyle \frac{\exp(x_a^{\top} \phi_{\mu})}{\sum\limits_{i=1}^K \exp(x_i^{\top} \phi_{\mu})}$
and
$\pi(a|x) = \displaystyle \frac{\exp(x_a^{\top} \phi_{\pi})}{\sum\limits_{i=1}^K \exp(x_i^{\top} \phi_{\pi})}.
$
The logging policy model $\hat{\mu}$ also follows the linear logistic model, while the value function model $\hat{Q}$ is defined by the linear regression model. Consequently, only the logging policy model is correctly specified.

\begin{figure}[!ht]
    \centering
    \includegraphics[width=0.4\textwidth]{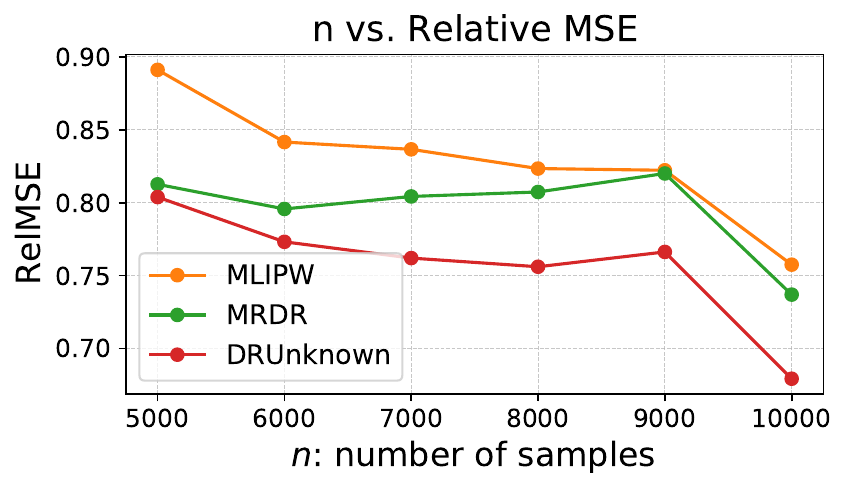}
        \caption{The relative MSE of the estimators on synthetic dataset with respect to the $n$, the number of samples.}
    \label{fig:n_vs_RelMSE}
\end{figure}

\begin{figure}[!t]
    \centering
    \includegraphics[width=0.43\textwidth]{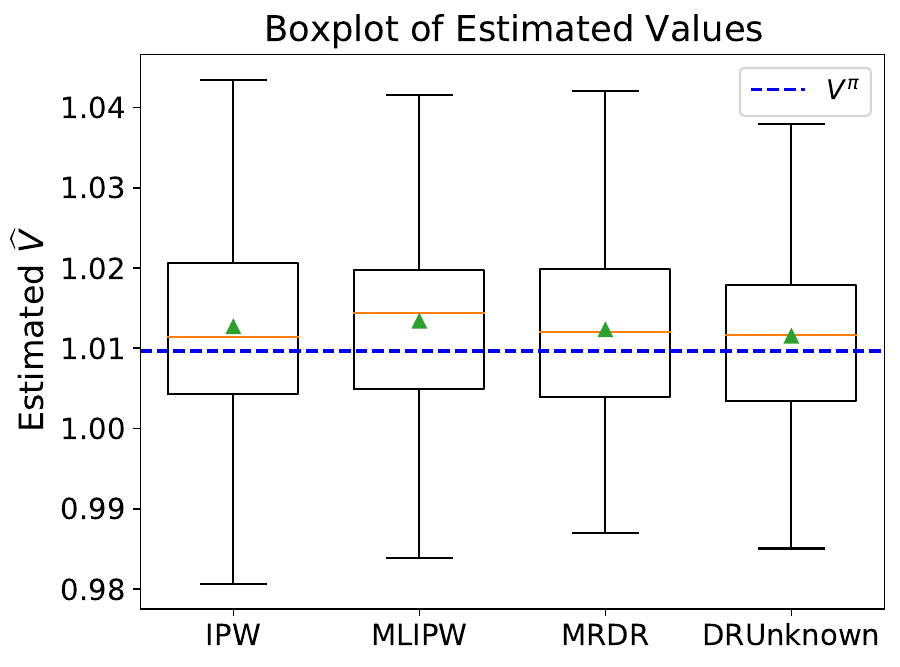}        
    \caption{Boxplot of estimated values from four estimators on simulation data with $N=100$ and $n=10000$. The true target policy value $V^{\pi}$ is indicated by the blue dashed line.}
    \label{fig:boxplot}
    \vspace{-2mm}
\end{figure}

The Table \ref{table:synthetic} and Figure \ref{fig:n_vs_RelMSE} reports the relative MSE values of MLIPW, MRDR and the proposed \EstName{}, calculated by dividing their MSE by the MSE of IPW, for the different sizes $n$ of each dataset. The proposed estimator achieves the lowest relative MSE.
Figure \ref{fig:boxplot} displays boxplots representing the estimated values of four methods, each computed with $N=100$ repeats and a dataset size of $n=10000$.

\begin{table}[!ht]
    \centering
    \begin{tabular}{c|ccc}
        sample size  & MLIPW  & MRDR   &  \textbf{\EstName{}}     \\ \hline
        $5000$ & 0.8911 & 0.8126 & \textbf{0.8038} \\
        $6000$ & 0.8416 & 0.7957 & \textbf{0.7731} \\
        $7000$ & 0.8366 & 0.8043 & \textbf{0.7620} \\
        $8000$ & 0.8234 & 0.8073 & \textbf{0.7560} \\
        $9000$ & 0.8222 & 0.8200 & \textbf{0.7662} \\
        $10000$ & 0.7575 & 0.7369 & \textbf{0.6792} \\
    \end{tabular}     
     \caption{The relative MSE of the estimators on synthetic dataset with respect to the $n$, the number of samples.}
     \label{table:synthetic}
\end{table}

\subsubsection{UCI Machine Learning Repository Dataset}\label{subsubsec:UCI}
For the real data experiment, we transform six classification datasets of the UCI Machine Learning Repository into contextual bandit problems : \texttt{glass}, \texttt{letter}, \texttt{zoo}\texttt{image}, \texttt{iris} and \texttt{handwritten} \citep{misc_glass_identification_42, misc_letter_recognition_59, misc_zoo_111, misc_image_segmentation_50, misc_iris_53, misc_optical_recognition_of_handwritten_digits_80}. Assigning the data to each class is considered as pulling an arm in the bandit. When the class is correct the reward is 1 and otherwise 0.

We construct the logging policy $\mu$ as follows: we train the policy $\mu_0$ using a linear logistic model on a separate dataset, and mix the policy with a random policy $\mu_{\text{random}}$ as $\mu=\alpha \mu_0 + (1-\alpha)\mu_{\text{random}}$, for $\alpha \in (0,1)$. The target policy $\pi$ is constructed as in the simulation data experiment.
The logging policy class is defined as $\{\muhat=\alpha \mu_0 + (1-\alpha)\mu_{\text{random}}:\alpha \in (0,1)\}$, where $\alpha$ serves as the parameter $\phi$, and we use the constant value function model for $\Qhat$.

\begin{figure}[!th]
    \centering
    \hspace{-11mm}
    \includegraphics[width=0.43\textwidth]{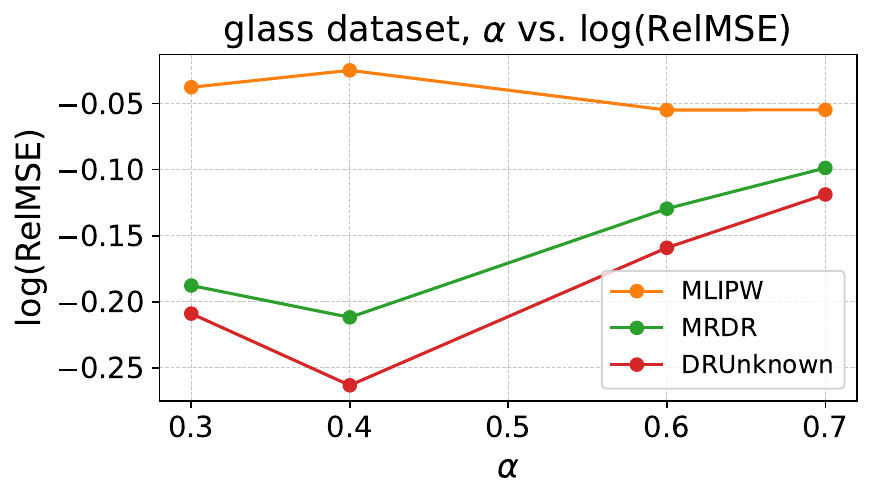}
    \vspace{-5mm}
    \caption{The logarithm of the relative MSE from three estimators on the \texttt{glass} dataset as a function of $\alpha$.}
    \label{fig:alpha_vs_RelMSE}
    \vspace{-5mm}
\end{figure}

We generate $N=100$ datasets, each with a size of $n=10000$, by randomly sampling contexts from the original dataset and selecting arms using the logging policy $\mu$. Table \ref{table:UCI} displays the relative MSE values of MLIPW, MRDR, and our proposed \EstName{}. Additionally, Figure \ref{fig:alpha_vs_RelMSE} depicts the log-relative MSE of each estimator on the \texttt{glass} dataset as a function of the logging policy parameter $\alpha$. The \EstName{} consistently demonstrates the smallest relative MSE across all six datasets.

\begin{table}[!t]
    \centering
    \begin{tabular}{c|ccc}
        dataset  & MLIPW  & MRDR   & \textbf{\EstName{}}        \\ \hline
        \texttt{glass} & 0.9548 & 0.8308 & \textbf{0.8125} \\
        \texttt{letter} & 0.9677 & 0.9323 & \textbf{0.9265} \\
        \texttt{zoo} & 0.9987 & 0.9842 & \textbf{0.9583} \\
        \texttt{image} & 0.8250 & 0.8245 & \textbf{0.8062} \\
        \texttt{iris} & 0.6116 & 0.5715 & \textbf{0.5454} \\
        \texttt{handwritten} & 0.9521 & 0.9432 & \textbf{0.9407} \\
    \end{tabular}
    \caption{The relative MSE of the estimators on UCI datasets, with $\alpha=0.4, n=10000$.}
    \label{table:UCI}
\end{table}

\subsection{Reinforcement Learning}
In this section, we provide the experimental results of OPE in the context of RL. We conducted the experiments on the ModelWin and ModelFail domains introduced in \citet{thomas2016data}. Detailed descriptions of these environments are available in the Appendix.
As done in Section \ref{subsubsec:UCI}, we utilize a mixture of the separately trained optimal policy and the uniform random policy with a rate $\alpha \in (0,1)$ as the logging policy $\mu$. We employ a linear logistic policy for the target policy $\pi$ and a linear model as the value function model $\Qhat$.

\begin{figure}[!t]
    \centering
    \includegraphics[width=0.43\textwidth]{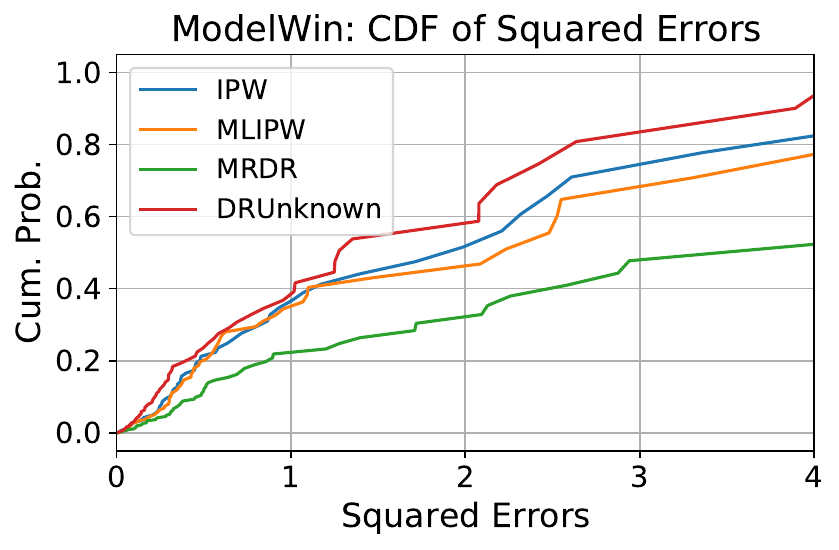}
    \vspace{-5mm}
    \caption{The CDF of squared errors for 4 estimators in ModelWin.}
    \label{fig:CDF}
\end{figure}

Table \ref{table:ModelWinFailComparison} displays the relative MSE of each estimator in ModelWin with $T=20$ and ModelFail environments. The proposed \EstName{} attains the lowest relative MSE in most scenarios, if the number of sample $n$ is large enough.
Figure \ref{fig:CDF} illustrates the cumulative distribution function (CDF) of squared error values (larger is better) on the ModelWin domain with $T=20$ and $n=20$. The \EstName{} exhibits the highest CDF values, indicating that the estimated values are more concentrated around the true value with high probability.

\subsection{Conclusion}
In conclusion, our study has introduced $\EstName{}$, a novel doubly-robust off-policy evaluation estimator for contextual bandits and reinforcement learning problems where both the logging policy and the value function remain unknown. Through a two-step estimation process, \EstName{} first estimates the logging policy using maximum likelihood estimator and subsequently estimates the value function model, aiming to minimize the asymptotic variance of the estimator while considering the impact of the logging policy estimation.

When the logging policy model is correctly specified, \EstName{} achieves the smallest asymptotic variance within the class that encompasses existing OPE estimators. Furthermore, if the value function model is also correctly specified, \EstName{} attains optimality by reaching the semiparametric lower bound for asymptotic variance.

To show the effectiveness of \EstName{}, we conducted experiments in both contextual bandits and reinforcement learning settings. The empirical results demonstrate the superior performance of \EstName{} compared to existing methods.

\section{Broader Impact}
This paper addresses the off-policy evaluation problem, aiming to contribute to the field of Machine Learning. It includes theoretical contributions and numerical experimental results, with no apparent societal or ethical issues that require further discussion.

\begin{table}[!t]
    \centering
    \begin{tabular}{c|ccc}
        ModelWin  & MLIPW  & MRDR   & \textbf{\EstName{}}        \\ \hline
        $n=10$ & 1.1604 & 1.9963 & \textbf{0.8916} \\
        $n=20$ & 1.0187 & 2.5577 & \textbf{0.8273} \\
        $n=30$ & 1.1064 & 1.7349 & \textbf{0.8529} \\
        $n=40$ & 1.0676 & 1.5270 & \textbf{0.9117} \\
    \end{tabular}
    \vspace{0.2em}  
    \begin{tabular}{c|ccc}
        ModelFail  & MLIPW  & MRDR   & \textbf{\EstName{}}        \\ \hline
        $n=10$ & 0.7794 & \textbf{0.7963} & 0.8476 \\
        $n=20$ & 0.5608 & 0.5591 & \textbf{0.2951} \\
        $n=30$ & 0.5532 & 0.5605 & \textbf{0.3154} \\
        $n=40$ & 0.5273 & 0.5285 & \textbf{0.1451} \\
    \end{tabular}    
    \caption{The realtive MSE of the estimators on ModelWin and ModelFail environments.}
    \label{table:ModelWinFailComparison}
\end{table}

\bibliography{ref}
\bibliographystyle{unsrtnat}

\newpage
\appendix

\section{Missing Proofs}

\subsection{Proof of Proposition \ref{prop:influence}}
\begin{proof}
    By the Taylor expansion,
$$\Vhat^{\text{DR}}(\betahat, \phihat) = V(\beta^*) + \Expectation_{\mu}{[\partialfrac{\Vhat}{\beta}]^{\top}}(\betahat - \beta^*) + \Expectation_{\mu}{[\partialfrac{\Vhat}{\phi}]^{\top}}(\phihat-\phi) + o_p(n^{-1/2}),$$
where
$$
V(\beta) = \frac{1}{n} \sumsample \Bigl[ \frac{\pisample}{\musample}
(r^{(i)} - \Qhatsample(\beta)) + \Vhatsample(\beta) \Bigr]  
$$
and $\betahat \xrightarrow{p} \beta^*$ for some $\beta^*$.
As $\muhat$ is correctly specified, we have
$$
\Expectation_{\mu}{[\partialfrac{\Vhat}{\beta}]} = \partialfrac{}{\beta}\Expectation_{\mu}{[-\frac{\pi(a|x)}{\mu(a|x)} \Qhat(x,a; \beta^*) + \Vhat(x; \beta^*)]} = 0.
$$
For the following term,
$$
-\Expectation_{\mu}{[\partialfrac{\Vhat}{\phi}]} = -\Expectation_{\mu}{[-\frac{\pi \dot{\muhat}}{\mu^2} (r - \Qhat(x,a;\beta^*)]} = \Expectation_{\pi}{[-\frac{ \dot{\muhat}}{\mu} (r - \Qhat(x,a;\beta^*)]} := \Gamma(\beta).
$$

For $\phihat-\phi$, from $U_n(\phihat)=0$ we have
$0 = U(\phi) + \Expectation{[\dot{U}(\phi)]}(\phihat-\phi) + o_p(n^{-1/2}),$
where 
$$
\Expectation{[\dot{U}(\phi)]} = \displaystyle\sumaction \Expectation_{\mu}{[\frac{1}{\mu(a|x)^2} \dot{\muhat}(a|x;\phi) \dot{\muhat}(a|x;\phi)^{\top}]} := \Sigma_{\phi \phi}
$$
and
$$
\phihat-\phi = -\Sigma_{\phi \phi}^{-1} U(\phi) + o_p(n^{-1/2}).
$$
Plugging in the results above and denoting $c(\beta)=\Sigma_{\phi \phi}^{-1} \Gamma(\beta)$, we have
\begin{equation*}
\begin{split}
\Vhat^{\text{DR}}(\betahat,\phihat) &= V(\beta^*) + \Gamma(\beta^*)^{\top}\Sigma_{\phi \phi}^{-1}U(\phi) + o_p(n^{-1/2}) \\
&= V(\beta^*) + c(\beta^*)^{\top} U(\phi) + o_p(n^{-1/2}) \\
&=  \frac{1}{n} \sumsample \etasample(\beta^*,c(\beta^*)) + o_p(n^{-1/2}),
\end{split}
\end{equation*}
for
$\etasample(\beta,c)=\displaystyle\frac{1}{\mu(a^{(i)}|x^{(i)})}\bigl[\pi(a^{(i)}|x^{(i)})r^{(i)}
-F(x,a;\beta,c,\phi) \bigr]  +\sumaction F(x^{(i)},a; \beta, c,\phi)$,
since $\sumaction \dot{\muhat}(a|x;\phi)=0.$
\end{proof}

\subsection{Proof of Proposition \ref{prop:bandit_variance}}
\begin{proof}
By the law of total variance, we decompose the variance of $\Vtilde(\beta,c)$ with $n=1$, given contexts $x$ and rewards $r$:
$$
\Var{(\Vtilde(\beta,c))} = \Var\Expectation{(\Vtilde(\beta,c))|x,r)} + \Expectation\Var{(\Vtilde(\beta,c))|x,r)}.
$$
For the first term, we have
$\Expectation_{\mu}{(\Vtilde(\beta,c))|x,r)} = V^{\pi}(x),$
and for the second term,
\begin{equation*}
\begin{split}
& \Var{(\Vtilde(\beta,c))|x,r)} = \Var_{\mu}\bigl[\frac{1}{\mu(a|x)}(\pi(a|x)r - F(x,a;\beta,c,\phi)) \bigr | x,r]\\
&= \Expectation_{\mu}\bigl[\frac{1}{\mu(a|x)^2}(\pi(a|x)r - F(x,a;\beta,c,\phi))^2 \bigr | x,r] - \Expectation_{\mu}\bigl[\frac{1}{\mu(a|x)}(\pi(a|x)r - F(x,a;\beta,c,\phi)) \bigr | x,r]^2 \\
&= \sumaction\frac{1}{\mu(a|x)}(\pi(a|x)r - F(x,a;\beta,c,\phi))^2 - [\sumaction (\pi(a|x)r - F(x,a;\beta,c,\phi)) ]^2 \\
&= \displaystyle\norm{\vec{F}(x;\beta,c,\phi)-\pi\vec{r} }_{M_{\mu}}^2
\end{split}
\end{equation*}
and since $\Expectation{[\pi \vec{Q}(x) - \pi \vec{r}]} =0$, we have
\begin{equation*}
\begin{split}
n\Var{(\Vtilde(\beta,c))} &= \Var (V^{\pi}(x)) + \Expectation_{\mu}\displaystyle\norm{\vec{F}(x;\beta,c,\phi)-\pi\vec{r} }_{M_{\mu}}^2 \\
&=\Var (V^{\pi}(x)) + \Expectation_{\mu}\displaystyle\norm{\vec{F}(x;\beta,c,\phi)-\pi\vec{Q}(x) }_{M_{\mu}}^2 + \Expectation_{\mu}\displaystyle\norm{\pi\vec{Q}(x)- \pi\vec{r}}_{M_{\mu}}^2.
\end{split}
\end{equation*}

\end{proof}

\subsection{Proof of Proposition \ref{prop:influence_RL}}
\begin{proof}
    The proof is similar to that of Proposition \ref{prop:influence} for contextual bandits, with $T=1$.
    
    For each $t \in [T]$, applying the Taylor expansion, we have
$$\Vhat_t(\betahat,\phihat) = V_t(\beta^*) + \Expectation_{\mu}{[\partialfrac{\Vhat_t}{\beta}]^{\top}}(\betahat - \beta^*) + \Expectation_{\mu}{[\partialfrac{\Vhat_t}{\phi}]^{\top}}(\phihat-\phi) + o_p(n^{-1/2}),$$
where
$$
V_t(\beta) = \frac{1}{n} \sumsample \rhosample_{0:{t-1}} \bigl[\rhosample_t [r_t^{(i)} - \Qhatsample_t(\beta)] + \Vhatsample_t(\beta)\bigr]
$$
and $\betahat \xrightarrow{p} \beta^*$ for some $\beta^*$.
As $\muhat$ is correctly specified, we have
$$
\Expectation_{\mu}{[\partialfrac{\Vhat_t}{\beta}]} = \partialfrac{}{\beta}\displaystyle\Expectation_{\mu}{[\rho_{0:t-1}[-\frac{\pi(a_t|x_t)}{\mu(a_t|x_t)} \Qhat(x_t,a_t; \beta^*) + \Vhat(x_t; \beta^*)]]} = 0.
$$
For the following term,
\begin{equation*}
\begin{split}
\Expectation_{\mu}{[\partialfrac{\Vhat_t}{\phi}]} &= \displaystyle\Expectation_{\mu}{\bigl[\partialfrac{\rho_{0:t-1}}{\phi}[\rho_t(r_t - \Qhat_t(x_t,a_t;\beta^*))+\Vhat_t(x_t,\beta^*)] + \rho_{0:t-1} \partialfrac{\rho_t}{\phi}(r_t-\Qhat_t(x_t,a_t;\beta^*))\bigr]}\\
&=\Expectation_{\mu}\bigl[ \partialfrac{\rho_{0:t-1}}{\phi} \Expectation_{\mu}[\rho_t(r_t - \Qhat_t(x_t,a_t;\beta^*))+\Vhat_t(x_t,\beta^*)|\History{t-1}] \bigr] \\
&+ \Expectation_{\mu} \bigl[\rho_{0:t-1} \partialfrac{\rho_t}{\phi}(r_t-\Qhat_t(x_t,a_t;\beta^*))\bigr] \\
&= \Expectation_{\mu} \rho_{0:t-1} \Expectation_{\mu}  \bigl[  \partialfrac{\rho_t}{\phi}(r_t-\Qhat_t(x_t,a_t;\beta^*)) | \History{t-1} \bigr] \text{ (the first term equals to zero)} \\
&= -\Expectation_{\mu} \rho_{0:t-1} \Expectation_{\pi} \bigl[ \frac{\dot{\hat{\mu}}(a_t|x_t;\phi)}{\mu(a_t|x_t)}(Q^t(x_t,a_t)-\Qhat^t(x_t,a_t;\beta^*))  \bigr] :=\Gamma_t(\beta^*).
\end{split}
\end{equation*}

For $\phihat-\phi$, from $U_n(\phihat)=0$ we have
$0 = U(\phi) + \Expectation{[\dot{U}(\phi)]}(\phihat-\phi) + o_p(n^{-1/2}),$
where 
$$
\displaystyle\sum\limits_{t=0}^{T-1} \sum\limits_{a \in \Aspace} \Expectation_{\mu}{[\frac{1}{\mu(a|x_t)^2} \dot{\hat{\mu}}(a|x_t;\phi) \dot{\hat{\mu}}(a|x_t;\phi)^{\top}]} := \Sigma_{\phi \phi}
$$
and
$$
\phihat-\phi = -\Sigma_{\phi \phi}^{-1} U(\phi) + o_p(n^{-1/2}).
$$
\begin{equation*}
\Vhat_t(\betahat,\phihat) = V_t(\beta^*) + \Gamma_t(\beta^*)^{\top} \Sigma_{\phi\phi}^{-1}U_n(\phi) + o_p(n^{-1/2}).
\end{equation*}
Plugging in the results above and denoting $c(\beta)=\displaystyle\sumtime \gamma^t \Sigma_{\phi \phi}^{-1} \Gamma_t(\beta)$, we have
\begin{equation*}
\begin{split}
\Vhat^{\text{DR}}(\betahat,\phihat) =\sumtime \gamma^t  \Vhat_t(\betahat,\phihat) &= V(\beta^*) + c(\beta^*)^{\top} U_n(\phi) + o_p(n^{-1/2}) \\
&=  \frac{1}{n} \sumsample \etasample(\beta^*,c(\beta^*)) + o_p(n^{-1/2}),
\end{split}
\end{equation*}
for
$V(\beta)=\sumtime \gamma^t V_t(\beta)$ and $\etasample(\beta,c) = \displaystyle\sumtime \gamma^t \etasample_t(\beta,c)$ with
$$
\eta_t^{(i)}(\beta,c) = \frac{1}{\mu(a_t^{(i)}|x_t^{(i)})}\bigl[ \rho_{0:t-1}^{(i)}\pi(a_t^{(i)}|x_t^{(i)})r_t^{(i)}
-F_{t}(x_t^{(i)},a_t^{(i)};\beta,c,\phi) \bigr]+\sum\limits_{a \in \Aspace} F_{t}(x_t^{(i)},a; \beta, c,\phi)
$$

\end{proof}

\subsection{Proof of Proposition \ref{prop:RL_variance}}
\begin{proof}
Section 4.1 of \cite{jiang2016doubly} introduces an inductive definition for a DR OPE estimator with a known logging policy and a fixed value function model. The Theorem 1 from the same paper provides the variance of this DR OPE estimator, presented as follows:>
\begin{equation*}
\begin{split}
n\Var[\Vhat] =\sumtime \gamma^{2t} \bigl[  \Expectation_{\mu} \bigl[ \Var_t \left[ \rho_{0:t-1}V^t(x_t) \right] + \Var_{t+1}[\rho_{0:t}r_t] \bigr] +  \Expectation_{\mu} \bigl[\Var_t [\rho_{0:t}(Q^t(x_t,a_t)-\Qhat(x_t,a_t))|x_t] \bigr] \bigr],
\end{split}
\end{equation*}
where $\Var_t$ refers to the conditional variance $\Var(\cdot|x_0,a_0, \dots, x_{t-1},a_{t-1})$.
The proof of the theorem still holds for a more general class of $\widehat{Q}$, which takes the state-action trajectory $(x_0,a_0, \dots, x_t,a_t)$ and the time step $t$ as arguments. Therefore, following the same approach as the proof of Theorem 1, we obtain the following representation of our proposed \EstName{} for RL:
\begin{equation*}
\begin{split}
n\Var[\Vhat^{\text{DR}}] &=\sumtime \gamma^{2t} \bigl[  \Expectation_{\mu} \bigl[ \Var_t \left[ \rho_{0:t-1}V^t(x_t) \right]+ \Var_{t+1}[\rho_{0:t}r_t] \bigr]\\
&+ \Expectation_{\mu} \bigl[\Var_t [\rho_{0:t}Q^t(x_t,a_t) - \mu(a_t|x_t)^{-1} F_{t}(x_t,a_t,\beta,c,\phi)|x_t]  \bigr] \bigr].
\end{split}
\end{equation*}
The first term does not depend on the parameters $\beta$ and $c$, so we denote it as a constant
$$C_T = \sumtime \gamma^{2t} \Expectation \bigl[ \Var_t \left[ \rho_{0:t-1}V^t(x_t) \right] + \Var_{t+1}[\rho_{0:t}r_t] \bigr].$$
The variance in the second term is conditioned on $(x_0,a_0,\dots,x_{t-1},a_{t-1},x_t)$, and thus the randomness of this variable only incurs from the action $a_t$ at time step $t$. Therefore, it can be calculated similarly to the proof of Proposition \ref{prop:bandit_variance}, and can be represented as a stochastic semi-norm as below, and we have the desired result.
\begin{equation*}
\begin{split}
&\Var_t [\rho_{0:t}Q^t(x_t,a_t) - \mu(a_t|x_t)^{-1} F_{t}(x_t,a_t,\beta,c,\phi)|x_t]\\
&= \Var_t [\sumaction \Delta_a^t [\rho_{0:t-1} \frac{\pi(a|x_t)}{\mu(a|x_t)} Q^t(x_t,a) - \mu(a|x_t)^{-1} F_{t}(x_t,\beta,c,\phi)|x_t]]\\ 
&= \norm{\vec{F}_{t}(x_t;\beta,c,\phi)-  \rho_{0:t-1} \pi \vec{Q}^t(x_t)}^2_{M_{t,\mu}}
\end{split}
\end{equation*}

\end{proof}

\subsection{Proof of Proposition \ref{prop:local_efficiency}}
\begin{proof}
Theorem 3 of \citet{kallus2020double} states that in the OPE problem, the DR estimator with the true logging policy $\mu$ and value function $Q$, given by
$$
\Vhat^{\text{opt}} = \frac{1}{n} \sumsample \sumtime \gamma^{t} \rho_{0:{t-1}}^{(i)}
\bigl[\rho_t^{(i)}[r_t^{(i)} - Q^t(x_t^{(i)}, a_t^{(i)})] + V^t(x_t^{(i)})\bigr],   
$$
achieves the smallest asymptotic variance, reaching the semiparametric lower bound, for the case with a discount factor $\gamma=1$.
When the value function model is also correctly specified, the proposed \EstName{} is asymptotically equivalent to $\Vhat^{\text{opt}}$, as the $\Gamma_t(\beta^*)=0$ for all $t$ and $c(\beta^*)=0$. 
For the general problem with $\gamma<1$, we can modify the MDP by fixing the discount factor to 1, incorporating the time step $t$ into the state variable $x$, and changing the reward function $R(x,a)$ to $\gamma^t R(x,a)$.
\end{proof}

\subsection{Proof of Proposition \ref{prop:intrinsic_efficiency}}
\begin{proof}
The IPW \cite{horvitz1952generalization}, DR \citep{dudik2011doubly,jiang2016doubly}, and MRDR \citep{farajtabar2018more} estimators are originally designed for the OPE problem with a known logging policy $\mu$. Therefore, for comparison, we assume that the logging policy model is estimated by MLE, as same as in our proposed estimator. We can observe that all three estimators are asymptotically equivalent to $\widetilde{V}(\beta,c)$ for some value of $\beta$ and fixed $c=0$. As all three estimators cannot take the estimation effect of $\hat{\phi}$ into account, they all have $c=0$.

The IPW sets the value of $\beta$ such that the value function model $\Qhat$ becomes zero, the DR estimator finds $\beta$ by minimizing the least-squares error for the value function. MRDR minimizes the variance of the estimator only with respect to $\beta$, with $c$ fixed to zero. The class of estimators contains all these estimators, and the proposed \EstName{} achieves the smallest asymptotic variance among them, being at least more efficient.
\end{proof}
\section{Estimation of $\phihat$ with General Estimating Equation}
The maximum likelihood estimator used to estimate $\hat{\phi}$ in this work is most efficient in many situations. However, the theoretical results in this paper are applicable to other estimating equations as given below,
\begin{equation*}
 U_n(\phi)=\sumsample \sumtime \sumaction (\Delta_{a,t}^{(i)} -\mu(a|x_t^{(i)}) )h(x_t^{(i)},a;\phi)=0, 
\end{equation*}
for any smooth function $h$. The equation $U_n(\phi)=0$ is an unbiased estimating equation as long as the logging policy model is correctly specified. A possible choice is to assign more weight to the state-action pair with high probability in $\pi$. This estimator remains consistent, satisfies the theoretical properties, and may be particularly useful for scenarios with a small-sized finite sample.

\section{Descriptions on Experimental Settings}
\subsection{Simulation Data}
To build the synthetic dataset for the simulation experiment, we generate elements for the context vectors $x$ and the coefficient vector $\beta$ from the uniform distribution $U(-1/\sqrt{d},1/\sqrt{d})$. The reward mean is determined by the nonlinear function $\exp(x^{t}\beta)$, making the linear value function model incorrectly specified.

For the logging policy $\mu$ and the target policy $\pi$, we generate the coefficients $\phi_{\mu}$ and $\phi_{\pi}$ from the uniform distribution $U(-1/\sqrt{d},1/\sqrt{d})$ and $U(-2/\sqrt{d},2/\sqrt{d})$, respectively.

\subsection{UCI Dataset}
The six datasets used in these experiments were initially designed for the classification problem. To transform the problem into a bandit setting, we interpret the assignment of the class label as the selection of an arm in a bandit, with a reward of 1 if the class is correct and 0 if incorrect. We train the classifier $\mu_0$, which returns the probability of each class label given a context vector $x$. We treat $\mu_0$ as a policy and combine it with a uniform random policy with rates $\alpha$ and $1-\alpha$. The value function model is constant and can be regarded as an intercept value.

\subsection{ModelWin and ModelFail}
\subsubsection{ModelWin}
ModelWin consists of three states, starting from state $s_1$. When choosing action $a_1$, the agent moves to $s_2$ with a probability of 0.6 and to $s_3$ with a probability of 0.4. Conversely, action $a_2$ leads to $s_2$ with a probability of 0.4 and to $s_3$ with a probability of 0.6. In states $s_2$ and $s_3$, both actions return the agent to $s_1$ with a probability of 1. If the agent visits $s_2$ or $s_3$, it receives rewards of 1 and -1, respectively. The horizon is fixed at $T = 20$.

\subsubsection{ModelFail}
ModelFail is an MDP with 4 states, but the learner cannot observe the current state of the agent. Starting from $s_1$, action $a_1$ leads to the upper middle state $s_2$, and $a_2$ to the lower middle state $s_3$. From both, any action moves to the terminal state $s_4$. If the transition is from the upper state, a reward of 1 is received; otherwise, a reward of -1. The horizon is always $T = 2$.

For both environments, the target policy at $s_1$ selects action $a_1$ with a probability of 0.7 and $a_2$ with a probability of 0.3. The logging policy chooses action $a_1$ with a probability of 0.75 and $a_2$ with a probability of 0.25. The parameter of the logging policy model $\muhat$ for both environments is the probability of choosing $a_1$, and the value function model $\Qhat$ is given by the linear model with intercepts.

\section{Limiations}
This paper primarily focuses on the asymptotic properties of the proposed estimator for the OPE problem with an unknown logging policy. We do not extensively explore the estimator's behavior with a finite sample. This aligns with similar statistical works, such as \cite{cao2009improving}, which addresses estimated missing mechanisms without providing finite-sample theory. Additionally, current DR OPE methods do not address scenarios with an unknown logging policy.

\end{document}